\pdfoutput=1
\documentclass{article}

\usepackage[nonatbib, preprint]{neurips_2021}

\usepackage[utf8]{inputenc} 
\usepackage[T1]{fontenc}    
\usepackage{hyperref}       
\usepackage{url}            
\usepackage{booktabs}       
\usepackage{amsfonts}       
\usepackage{nicefrac}       
\usepackage{microtype}      
\usepackage{xcolor}         

\usepackage{epsfig}
\usepackage{graphicx}
\usepackage{amsmath}
\usepackage{amssymb}

\usepackage[abbreviations]{glossaries-extra}
\glsdisablehyper

\glssetcategoryattribute{acronym}{indexonlyfirst}{true}

\newabbreviation{cnn}{CNN}{Convolutional Neural Network}
\newabbreviation{rnn}{RNN}{Recurrent Neural Network}
\newabbreviation{mlp}{MLP}{Multilayer Perceptron}
\newabbreviation{nerf}{NeRF}{Neural Radiance Field}
\newabbreviation{gan}{GAN}{Generative Adversarial Network}
\newabbreviation{dip}{DIP}{Deep Image Prior}
\newabbreviation{mse}{MSE}{Mean Squared Error}
\newabbreviation{snr}{SNR}{Signal-to-Noise Ratio}

\makeglossaries

\title{Neural Knitworks:\\Patched Neural Implicit Representation Networks}

\author{%
  Mikolaj Czerkawski \And Javier Cardona \And Robert Atkinson \And Craig Michie \And Ivan Andonovic \And Carmine Clemente \And Christos Tachtatzis
  \AND\\
  University of Strathclyde
}

\begin{document}

\maketitle

\begin{abstract}

    Coordinate-based Multilayer Perceptron (MLP) networks, despite being capable of learning neural implicit representations, are not performant for internal image synthesis applications. Convolutional Neural Networks (CNNs) are typically used instead for a variety of internal generative tasks, at the cost of a larger model. We propose \textit{Neural Knitwork}, an architecture for neural implicit representation learning of natural images that achieves image synthesis by optimizing the distribution of image patches in an adversarial manner and by enforcing consistency between the patch predictions. To the best of our knowledge, this is the first implementation of a coordinate-based MLP tailored for synthesis tasks such as image inpainting, super-resolution, and denoising. We demonstrate the utility of the proposed technique by training on these three tasks. The results show that modeling natural images using patches, rather than pixels, produces results of higher fidelity. The resulting model requires 80\% fewer parameters than alternative CNN-based solutions while achieving comparable performance and training time.
    
\end{abstract}

\section{Introduction}

    The research on utilizing coordinate-based \gls{mlp} networks for image synthesis has developed significantly, yielding a range of impressive results~\cite{fourier_features, siren, Mildenhall2020, Zhang2020, Yu2020, Niemeyer2020}. However, most of the published works propose architectures that have no capability to model directly the spatial relationships within the represented signal, they rather attempt to independently fit network output for a set of input coordinates. In this work, we propose a coordinate-based model that has spatial awareness by fitting patches to input coordinates rather than isolated values.
    
\begin{figure*}
        \centering
        \includegraphics[width=\textwidth]{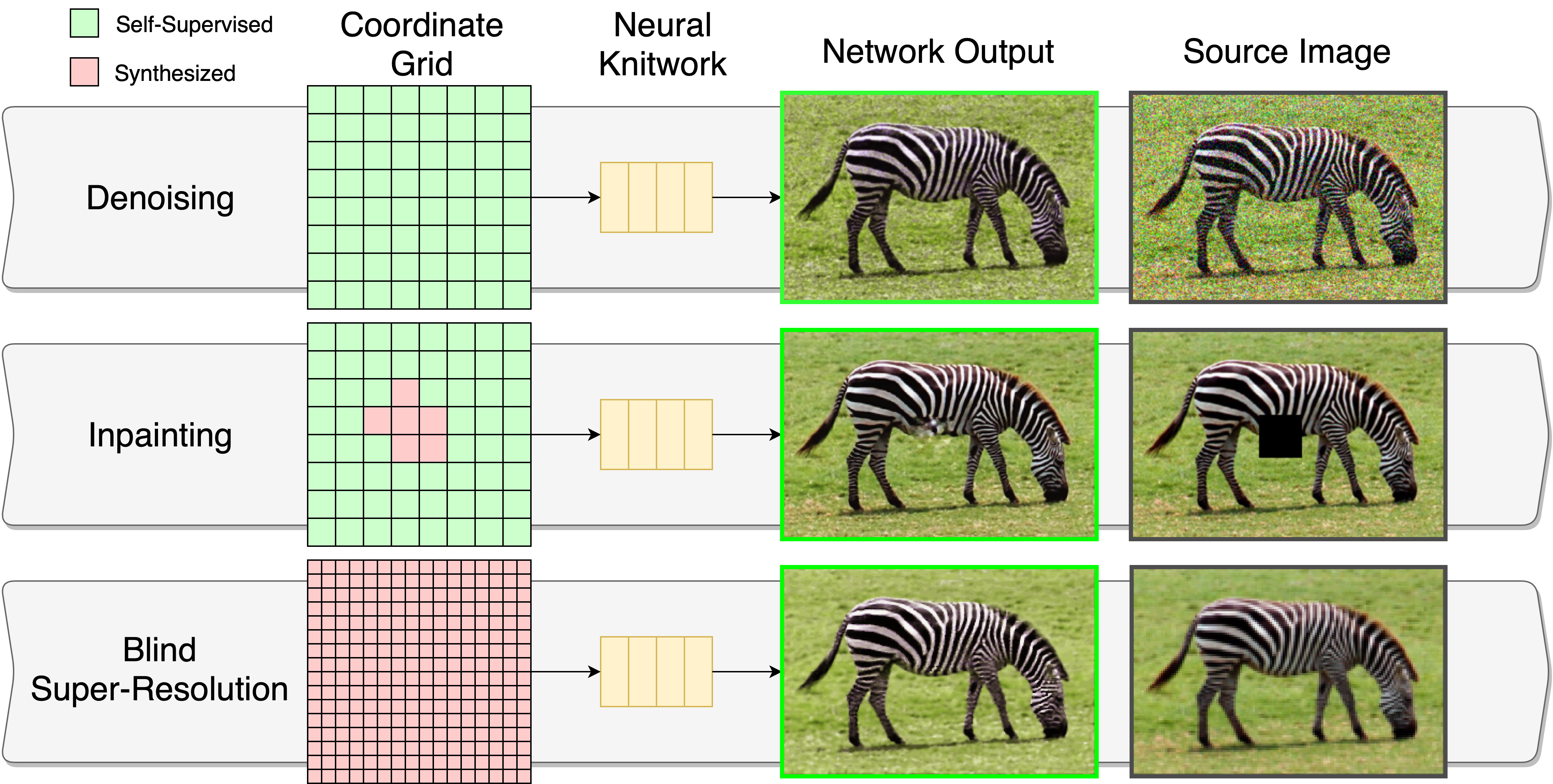}
        \caption{The introduced model trained on a single sample can perform a number of different image synthesis tasks with very low memory requirements.}
        \label{fig:intro}
    \end{figure*}

    The idea is inspired by the advancements made using models that focus on patch distributions like InGAN~\cite{Shocher2019}, SinGAN~\cite{Shaham2019}, and the Swapping Autoencoder~\cite{Park2020}. The proposed framework is an improvement to the conventional coordinate \gls{mlp} architectures, where the network predicts a color patch (or a multi-scale stack thereof) with additional constraints imposed. The purpose of these constraints is to match the distributions of predicted and reference patches and encourage spatial consistency between the predictions. The resulting method constitutes a framework that can be applied to several image synthesis tasks, such as image inpainting, super-resolution and denoising, as shown in Figure~\ref{fig:intro}.
    
    The proposed approach combines the advantages of \gls{mlp}s as neural implicit representation networks while providing versatility and robustness in levels similar to a \gls{cnn}. The network can be significantly smaller than equivalent \gls{cnn} architectures and faithfully encode a target signal, exhibiting a compressive capability~\cite{Dupont2021}. Furthermore, fitting a single image by the network has two significant advantages: i) there is no requirement for a dataset or pretraining ii) it requires fewer iterations to converge compared to solutions trained on a dataset of images. Effectively, a flexible internal learning framework is introduced that performs well on a diverse range of computer vision tasks with low memory and computational requirements.
    
\section{Related Work}

    The potential of applying a \gls{mlp} network as an encoding of a signal has recently been explored in a number of works~\cite{fourier_features, siren, Mildenhall2020, Zhang2020, Niemeyer2020, Dupont2021, Skorokhodov2020, Anokhin2020, Park2019, Genova2019, Atzmon2019}. The learned signals can be of any dimensionality, however, \gls{mlp} encoding of spatial coordinates is a particularly popular theme, involving a network that learns to produce given scalar values based on the input coordinates. This allows for considerable flexibility and leads to applications such as self-supervised learning of natural images or videos.

    \textbf{Coordinate-Based \gls{mlp} Networks.}
    The interest in using fully connected networks to represent signals in an implicit manner has grown over the last few years, which can be attributed to the potential of such methods to be used for 3D shape representations~\cite{Mildenhall2020, Zhang2020, Niemeyer2020,Park2019, Genova2019,Atzmon2019}.
    An important issue for learning coordinate-based representations is the tendency of neural networks to interpolate and attenuate high-frequency changes in the output~\cite{fourier_features,siren,spectral_bias}. Two effective solutions to this problem are to either map the input coordinates (known as positional encoding)~\cite{fourier_features} or use sinusoidal activation functions~\cite{siren}. However, neither of the two approaches does address the challenge of synthesizing new regions. As we demonstrate in a subsequent section (Figure~\ref{fig:ablation}), a standard \gls{mlp} encoding input with random Fourier features does not synthesize new outputs in a convincing manner.
    The novel techniques of random Fourier feature encoding of spatial coordinates gave rise to \gls{nerf} networks, which can synthesize high fidelity novel views of 3D scenes in an efficient manner~\cite{Mildenhall2020}. This contribution was soon followed by further developing works, focusing on aspects such as unbounded 3D scenes~\cite{Zhang2020}, synthesizing based on few (or only one) images \cite{Yu2020}, or taking advantage of compositionality of 3D scenes~\cite{Niemeyer2020}.
    There have been some works where coordinate-based \gls{mlp} networks are used as a core for a generative model using techniques such as a hypernetwork predicting the weights of a sample coordinate \gls{mlp}~\cite{Skorokhodov2020}, or by modulating the weights of a base coordinate \gls{mlp}~\cite{Anokhin2020}. These approaches are fundamentally different as they attempt to create a wide generative model based on a large-scale dataset, while our approach focuses on data-agnostic internal learning tasks and uses a disparate architecture. Finally, Local Implicit Image Functions introduced in~\cite{Chen2020} are trained in a self-supervised manner and are based on latent feature maps used to synthesize an image at different resolutions. However, the architecture relies on a convolutional feature encoder, applies a fixed downsampling operation, and is trained to generate images based on a selected dataset. Our architecture is purely based on \gls{mlp} networks, requires no pretraining, and directly maximizes self-similarity between the synthesized and known patches.
    \textbf{Internal Generative Frameworks.} Patches have been identified as crucial representation features of image in various works~\cite{Shaham2019, Efros2001, Kwatra2003, Simakov2008, Glasner2009, Zontak2011, Zontak2013, Gatys2015, Michaeli2014, Shocher2018, Gandelsman2019, Mechrez2019, Bell-Kligler2019}. The introduction of \gls{gan}s~\cite{Goodfellow2014} made it possible to learn patch distributions of images in an adversarial manner~\cite{Shaham2019, Park2020}. Additionally, internal learning approaches relying on the priors contained in convolutional architectures have been proposed~\cite{Gandelsman2019, Ulyanov2020}. To the best of our knowledge, no attempt of introducing these techniques to coordinate-based \gls{mlp} networks has been made until now.

\section{Method}

    \begin{figure*}
        \centering
        \includegraphics[width=\textwidth]{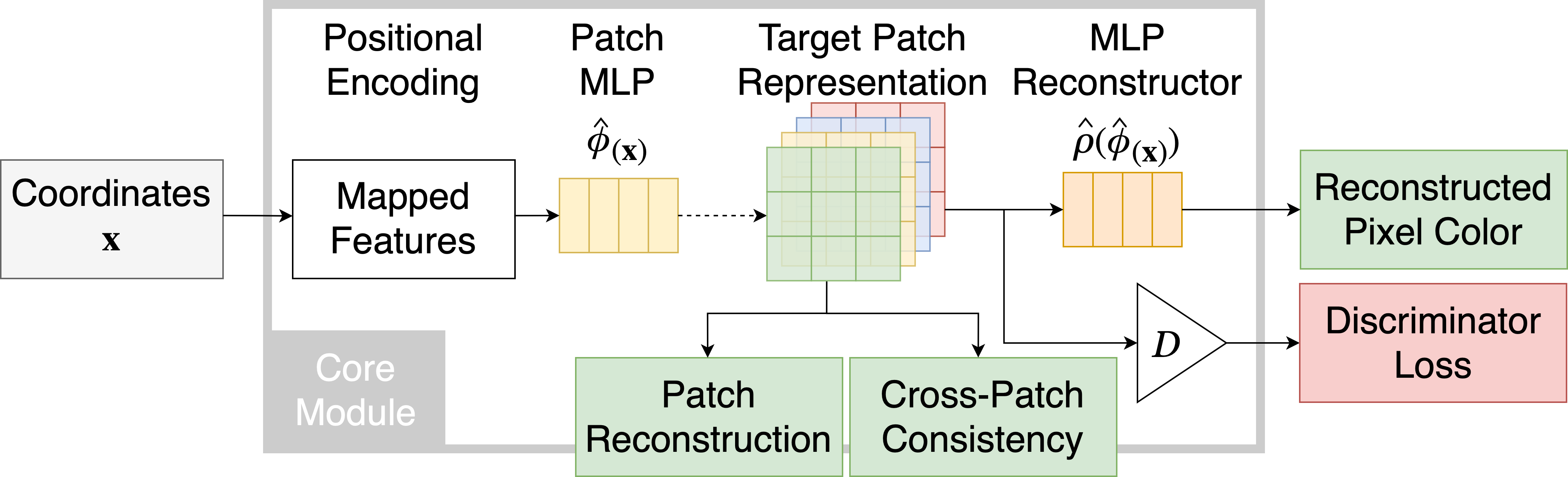}
        \caption{Neural Knitwork architecture consists of 3 shallow \gls{mlp}s. The network knits patches for related coordinates by enforcing consistency of predictions and optimizing likelihoods of individual patches. Each patch stack is translated back to a single color by the \gls{mlp} Reconstructor.}
        \label{fig:basic_diagram}
    \end{figure*}
    
    The core structure of the proposed network is presented in Figure~\ref{fig:basic_diagram}. It consists of three small networks: (i) Patch \gls{mlp} for translating from the original coordinate domain to the patch domain (ii) the discriminator responsible for assessing patch likelihoods, and (iii) \gls{mlp} Reconstructor for mapping the patch domain to individual pixel color.
    
    The resulting architecture performs the equivalent operation to a conventional coordinate-based \gls{mlp} since the network ultimately predicts a single pixel value. However, the intermediate patch-based representation of the proposed architecture forces the model to establish the natural relationship between the encoded coordinates. This property can also be used as a useful prior for internal learning scenarios, similar to using convolutional kernels in \gls{cnn} architectures. Further, the patch representation allows our model to be trained as a \gls{gan} and match the internal patch distribution with that of the reference image.
    
    \subsection{Patch Synthesis}
        \label{subsec:patch_output}
        
        The Patch \gls{mlp} is a network of 4 ReLU layers with 256 units, identical to the one used in~\cite{fourier_features}. The role of this component is to map each coordinate vector to an appropriate pixel patch. The coordinate input is mapped using random Fourier features before passing to the network. This processing step is known as positional encoding and has been described in detail in~\cite{fourier_features}.

        The output of this network approximates the implicit representation function $\phi_{(\textrm{\textbf{x}})}$ for a query coordinate vector $\textrm{\textbf{x}}$ along with values of neighbouring coordinates. The required receptive field depends on the spectral content of the image and can be adjusted by either increasing the patch size to provide more spatial bandwidth or using multi-scale patches. We apply the latter approach as it is more efficient for large spatial spans, allowing for easily configurable scope covered by the output patches at low computational cost. We use patches of fixed size 3 by 3 for all experiments. For extraction of the patches with scales larger than one, a Gaussian filter is applied to the image to reduce aliasing.

        \textbf{Patch Reconstruction Loss} Since our core module is an \gls{mlp} with multi-scale patch output, a direct way of computing the error is taking the difference of the predicted patches $\hat{\phi}_{(\textrm{\textbf{x}})}$ and ground truth reference $\phi_{(\textbf{x})}$. For inpainting tasks, not all pixel values for the patch stack are known and, hence, we apply an appropriate mask $\textrm{\textbf{m}}_{(\textbf{x})}$ to this loss. For other tasks, the mask will be a unit tensor. We refer to this loss as patch reconstruction loss $\mathcal{L}_{Recon}$, which is effectively a masked \gls{mse} computed for patches at $N$ sampling coordinates.
        
        \begin{equation}
            \mathcal{L}_{Recon} = \sum_{\textrm{\textbf{x}}}^{N} \frac{(\phi_{(\textrm{\textbf{x}})} - \hat{\phi}_{(\textrm{\textbf{x}})})^2*\textrm{\textbf{m}}_{(\textrm{\textbf{x}})}}{|\phi_{(\textrm{\textbf{x}})}|}
        \end{equation}
        
        The effect of learning patch-based representation rather than direct pixel values has been illustrated in Figure~\ref{fig:ablation} as part of the ablation study included in the experiments. It becomes quite clear that patch-based representation alone (third column), while helpful, may not yield satisfactory results for challenging synthesis tasks. Instead, we must apply additional constraints to control the relationships between the synthesized values.

        \textbf{Cross-Patch Consistency Loss} The ability to produce likely pixels or patches does not necessarily lead to consistent network output when the entire learned image is considered. By default, all patches for which ground truth is available, are optimized to be close to that reference, but this does not guarantee that all patches contribute to a single coherent image for coordinates with no ground truth. For new synthesized regions, the output patches may be convincing on their own (due to the bias component learned by the network from the known region) but display limited coherence between each other.
        
        To encourage consistency, we design a cross-patch consistency loss that computes the difference between predictions for each pixel from all patches and for the entire image scope. In practice, a way to enforce this, is to use the predictions from the central element of the lowest-scale patch as a reference. The following notation is defined: $\hat{\phi}_{(\textrm{\textbf{x}})}[\textrm{\textbf{i}}]$ represents the  value of a patch element $\textrm{\textbf{i}}$ predicted for coordinate $\textrm{\textbf{x}}$ where $\textrm{\textbf{i}}$ belongs to the the set of $I$ elements across all scales. In a similar fashion, the $\hat{\phi}_{(\textrm{\textbf{x}})}[\textrm{\textbf{o}}]$, represents the value of the central element (constant index of $\textrm{\textbf{o}}$) of the lowest scale patch predicted from coordinate $\textrm{\textbf{x}}$.
        
        The central reference $\hat{\phi}_{(\textrm{\textbf{x}})}[\textrm{\textbf{o}}]$ is compared with element $\hat{\phi}_{(\textrm{\textbf{x+s}})}[\textrm{\textbf{i}}]$ that corresponds to the same pixel of the output image evaluated at coordinates $\textrm{\textbf{x+s}}$, where $\textrm{\textbf{s}}$ indicates the appropriate shift, dependent on $\textrm{\textbf{i}}$. The terms with values $\textrm{\textbf{x+s}}$ outside of the image bounds are naturally excluded from the summation. 
        
        \begin{equation}
            \mathcal{L}_{\textrm{X-patch}}  = \sum_{\textrm{\textbf{x}}}^{N}\sum_\textrm{\textbf{i}}^I (\hat{\phi}_{(\textrm{\textbf{x+s}})}[\textrm{\textbf{i}}] - \hat{\phi}_{(\textrm{\textbf{x}})}[\textrm{\textbf{o}}])^2
        \end{equation}

        \textbf{Reconstructed Pixel Loss} The transition from predicting isolated pixel colors to patches introduces a new trade-off between imposing spatial relationships of the pixel colors and obtaining a high fidelity image with accurate detail. In practice, there will be some disagreement between the predictions for the same pixel from different patches and scales. The naive approach of averaging all predictions for a given coordinate value leads to blurring. To avoid this, a separate \gls{mlp} Reconstructor network is used to translate from a multi-scale patch representation to a single color value, by approximating the color extraction function $\rho((\hat{\phi}_{(\textrm{\textbf{x}})})$, as shown in Figure~\ref{fig:basic_diagram}. The error made by this final output network constitutes the reconstructed pixel loss, encouraging the entire model to produce accurate pixel colors based on a stack of patches.
        
        The pixel reconstruction loss is computed as a $\ell_1$ loss between the network pixel color output $\hat{\rho}(\hat{\phi}_{(\textrm{\textbf{x}})})$ and the color ground truth $\textrm{\textbf{c}}(\textrm{\textbf{x}})$
        
        \begin{equation}
            \mathcal{L}_{\textrm{Pixel}} =  \sum_{\textrm{\textbf{x}}}^{N} |\hat{\rho}(\hat{\phi}_{(\textrm{\textbf{x}})}) - \textrm{\textbf{c}}(\textrm{\textbf{x}})|
        \end{equation}

    \subsection{Patch Discriminator}
        Another important property to enforce, especially when some parts of the signal need to be synthesized, is for all predicted patches to come from a distribution of likely patches, derived from the available information in the source image. This is achieved with the aid of a discriminator tasked to predict which patches come from the original distribution and which do not. The approach is partly inspired by a number of existing works that take advantage of self-similarity between patches in natural images~\cite{Shaham2019, Park2020, Glasner2009,Shocher2018, Gandelsman2019}. In our case, the discriminator is another \gls{mlp} consisting of 3 Leaky ReLU layers and taking a flattened patch representation as input.
        
        \textbf{Discriminator Loss} The discriminator network takes a single multi-scale patch and outputs a confidence score. At each training step of the discriminator, we feed it all real and all synthesized patches and compute the output confidence for them. Furthermore, we apply one-sided label smoothing~\cite{Salimans2016} of the real labels with a factor of 0.1 when computing the discriminator loss in order to penalize over-confidence of this network module. We use a standard binary cross-entropy loss on the discrimination scores.
        
    \subsection{Complete Objective Function}
        The objective function is a minimax loss where the generator loss term is composed of the four losses contained parameterized by weights $\alpha$, $\beta$, and $\gamma$.
    \begin{equation}
        \label{eq:total_loss_g}
        \mathcal{L}_{Total, G} = \mathcal{L}_{Recon} + \alpha \mathcal{L}_{\textrm{X-patch}} + \beta \mathcal{L}_{\textrm{Pixel}} + \gamma \mathcal{L}_{\textrm{BCE,G}}
    \end{equation}
    The discriminator term only includes a single binary-cross entropy loss. Further details about the implementation and the hyperparameters can be found in the supplementary material.

\section{Experiments}

    We demonstrate the capabilities of Neural Knitworks by utilizing a similar model with only minor adjustments for several tasks commonly investigated in the field of computer vision: 1) image inpainting 2) super-resolution and 3) denoising. The following section describes the key implementation details for each task and presents corresponding qualitative results. Furthermore, quantitative measures are provided by applying each method to Set5~\cite{set5paper} and Set14~\cite{set14paper}.

    \subsection{Ablation Study}
    
    \begin{figure*}[t]
        \centering
        \includegraphics[width=\textwidth]{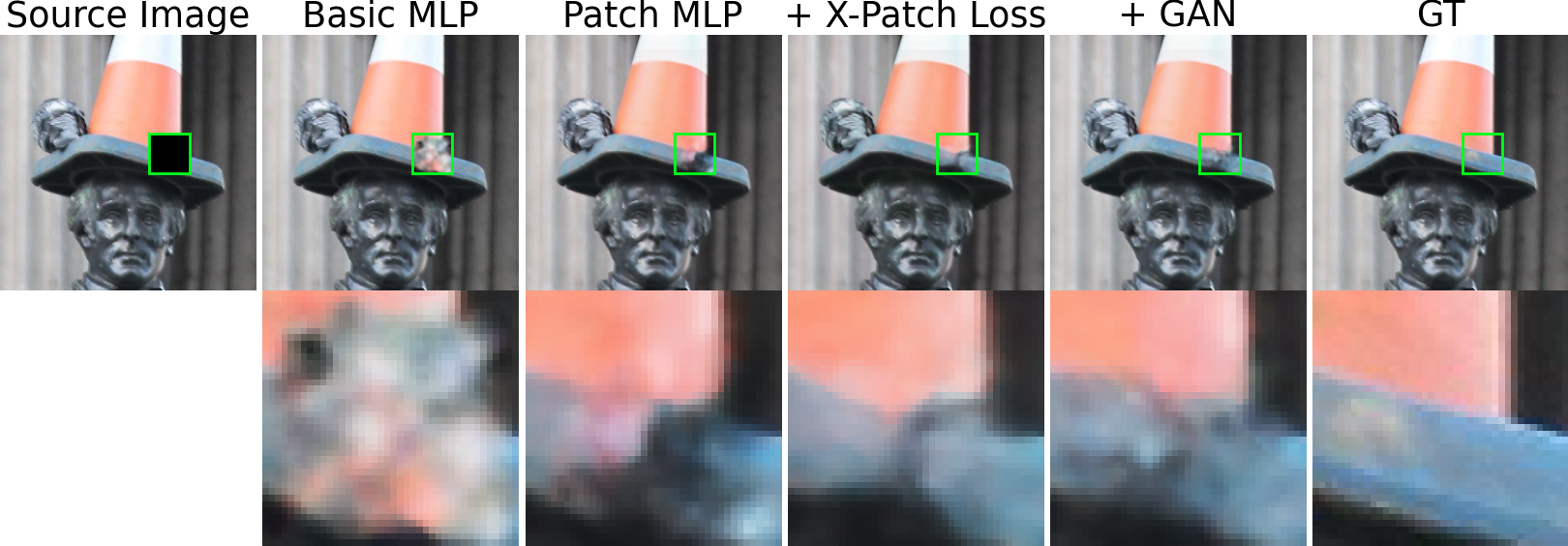}
        \caption{Ablation study of Neural Knitwork components. Conventional \gls{mlp} does not produce coherent inpainted region and this is improved with the introduction of patches. Further, imposing cross-patch consistency constraint increases the quality of the synthesized region while employing a \gls{gan} approach ensures patches of high likelihood.}
        \label{fig:ablation}
    \end{figure*}
    
    We begin our analysis with an ablation study of the proposed architecture to demonstrate the utility of each introduced loss component. Figure~\ref{fig:ablation} illustrates the effect of the following adjustments to the conventional coordinate \gls{mlp} network (second column): i) patch output (third column), ii) cross-patch consistency loss (fourth column), iii) patch discrimination (fifth column). We observe that the introduction of patch output alone can lead to a more convincing synthesis. However, some distortion can be observed in the synthesized region, which is reduced when cross-patch consistency loss is used. Finally, the addition of a \gls{gan} loss leads to improved region consistency.
    
    \begin{figure*}[h]
        \centering
        \includegraphics[width=0.8\textwidth]{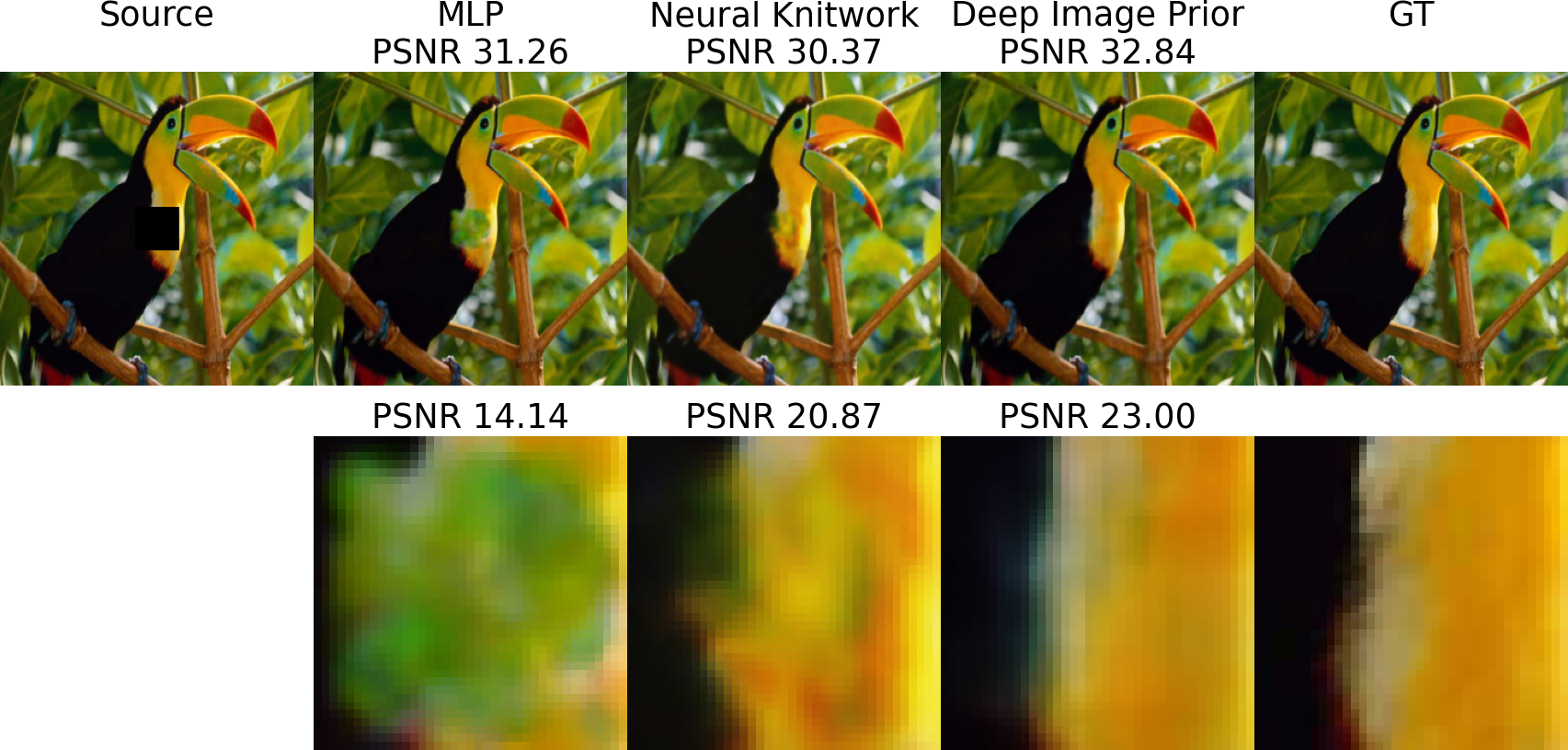}
        \includegraphics[width=0.8\textwidth]{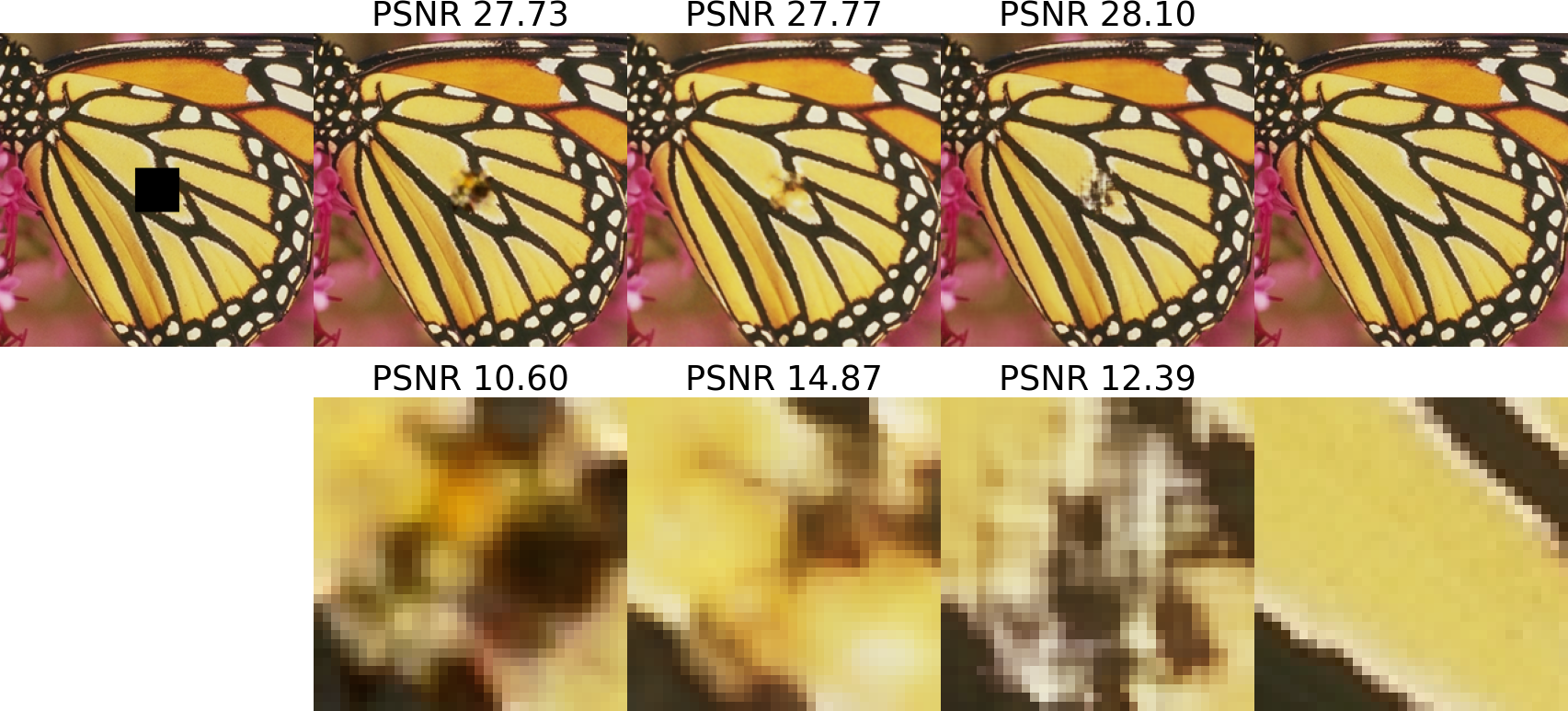}
        \caption{Image inpainting results for a fill ratio of 2\%. For the inpainted region Neural Knitworks and \gls{dip} perform comparably, and both outperform conventional \gls{mlp}.}
        \label{fig:inpainted_images}
    \end{figure*}
    
    \subsection{Image Inpainting}
        
    For the image inpainting task, we cut out a rectangular section from the source image to be used as the inpainted region. The coordinates of the cutout are used for producing a mask indicating whether the source signal exists for a given pixel. The mask is used to backpropagate the reconstruction losses only from the pixels outside the inpainted region.

    We compare the results of the inpainting for the Neural Knitwork to a conventional coordinate \gls{mlp} model and to \gls{dip}~\cite{Ulyanov2020}, a \gls{cnn}-based internal learning approach. Figure~\ref{fig:inpainted_images} contains the resulting output for the three tested models. 
    The reconstruction quality of the whole image is comparable for the three tested methods. However, when inpainted region is concerned, we observe a significant improvement of over 4 dB for the Neural Knitwork compared to the conventional coordinate \gls{mlp} and 2 dB less than the \gls{cnn}-based technique. For some of the results, the Neural Knitwork was, in fact, able to outperform \gls{dip}. Table~\ref{tab:inpainting_comparison} contains the evaluation across the entire datasets for different fill ratios, which supports that the Neural Knitwork outperforms the conventional approach and achieves comparable performance to the \gls{dip} with approximately 80\% less parameters. More examples can be found in the supplementary material.
    
    \begin{figure*}[h]
        \centering
        \includegraphics[width=\textwidth]{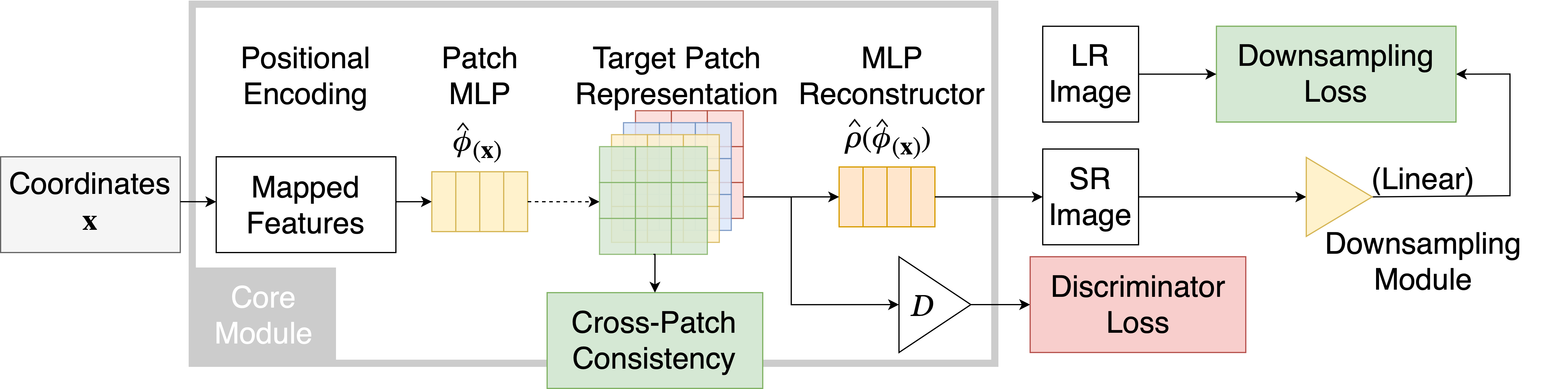}
        \caption{The blind super-resolution framework utilizes the core module with the addition of a linear network to blindly infer the downsampling kernel. In this case the patch reconstruction loss can not be computed.}
        \label{fig:sr_diagram}
    \end{figure*}

    \begin{table}[]
        \centering
        \caption{Comparison of inpainting performance for different fill ratios. The three approaches appear comparable PSNR ($\uparrow$) and SSIM ($\uparrow$) for \textit{whole images}. For the \textit{inpainted region}, the Neural Knitwork comes close to the level of performance of \gls{dip}, while conventional \gls{mlp} is inferior.}
        \begin{tabular}{ccccc}
        \hline
            Dataset & Fill Ratio & MLP & \gls{dip}~\cite{Ulyanov2020} & Neural Knitwork (ours) \\
            \hline
            \hline
             & 1\% & \textbf{32.99/0.98} & 32.53/0.95 & 32.00/0.96 \\
            Set5 & 2\% & 28.65/0.97 & 29.35/0.92 & \textbf{29.81/0.94} \\
            \textit{Whole Image} & 4\% & 25.85/0.96 & 26.22/0.88 & \textbf{27.96/0.95}\\
             \hline
             & 1\% & 13.96/0.36 & \textbf{20.66/0.68} & 18.28/0.58 \\
            Set5 & 2\% & 11.89/0.28 & \textbf{18.50/0.57} & 17.79/0.57 \\
            \textit{Inpainted Region} & 4\% & 11.89/0.32 & \textbf{15.89/0.52} & 14.95/0.48\\
            \hline
             & 1\% & \textbf{28.97/0.95} & 28.22/0.90 & 27.65/0.91 \\
            Set14 & 2\% & 26.38/0.94 & \textbf{27.08/0.89} & 26.03/0.91\\
            \textit{Whole Image}  & 4\% & 24.00/0.93 & \textbf{25.53/0.89} & 24.44/0.89\\
            \hline
             & 1\% & 11.85/0.23 & \textbf{16.32/0.41} & 15.50/0.40 \\
            Set14 & 2\% & 10.79/0.23 & \textbf{15.32/0.40} & 13.94/0.39\\
            \textit{Inpainted Region}  & 4\% & 10.67/0.24 & \textbf{14.08/0.37} & 12.35/0.36\\
            \hline
            \rule{0pt}{2.5ex} Parameters & & 263K & 2,400K & 512K \\
        \end{tabular}
        \label{tab:inpainting_comparison}
    \end{table}
    
    \subsection{Super-Resolution}
    
    To perform super-resolution, a Neural Knitwork has to translate the information contained in the patches of the original scale to a domain of patches of finer scale. This can be done by matching the patch distribution across scales~\cite{Shaham2019,Michaeli2014, Shocher2018,Bell-Kligler2019}. For blind super-resolution, Neural Knitwork core module is utilized with adjusted losses as illustrated in Figure~\ref{fig:sr_diagram}. The queried coordinates for a patch \gls{mlp} network include all super-resolved coordinates, which means that it is not possible to compute the patch reconstruction loss in this mode. However, it is possible to compute the cross-patch consistency loss as well as discriminate the patches to match the source image distribution. This alone could yield an output image resembling the low-resolution source without guaranteed structural coherence. To enforce coherence, we apply spatially-aware supervision by downsampling the super-resolved image and computing the downsampling loss with the reference to the low-resolution source image.
    
    The downsampling operation can be implemented in several ways. If the downsampling kernel is known, then the best approach is to simply backpropagate through that kernel (assuming it is differentiable). Otherwise, we can create a trainable downsampling module representing the kernel and optimize its weights in an end-to-end manner. We revisit the technique introduced in~\cite{Bell-Kligler2019} by using an identical deep linear network to approximate the kernel. Their method relies on the assumption that a satisfactory kernel should preserve the distribution of patches in the image. For Neural Knitworks, there is no need to introduce a new loss term accommodating this since the core module objective imposes matching patch distribution by default.
    
    \begin{figure*}
        \centering
        \includegraphics[width=0.9\textwidth]{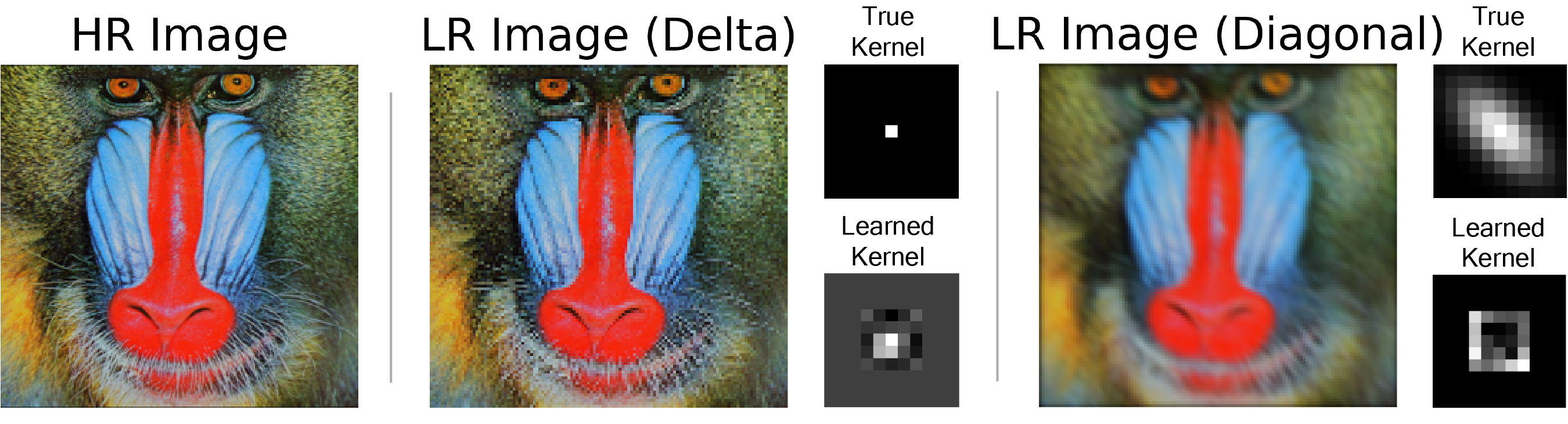}
        \caption{Our method approximates the downsampling kernel depending on the source image.}
        \label{fig:kernel_sr_images}
    \end{figure*}
    
    In Figure~\ref{fig:kernel_sr_images}, we demonstrate the downsampling effect of two non-standard kernels: i) delta function (leading to aliasing) and ii) diagonal Gaussian kernel. Different types of artifacts can be observed depending on the kernel. During training, Neural Knitwork blindly approximates the downsampling kernel based on the image content. The true and learned kernels are illustrated in the figure.
    
\begin{figure*}
        \centering
        \includegraphics[width=\textwidth]{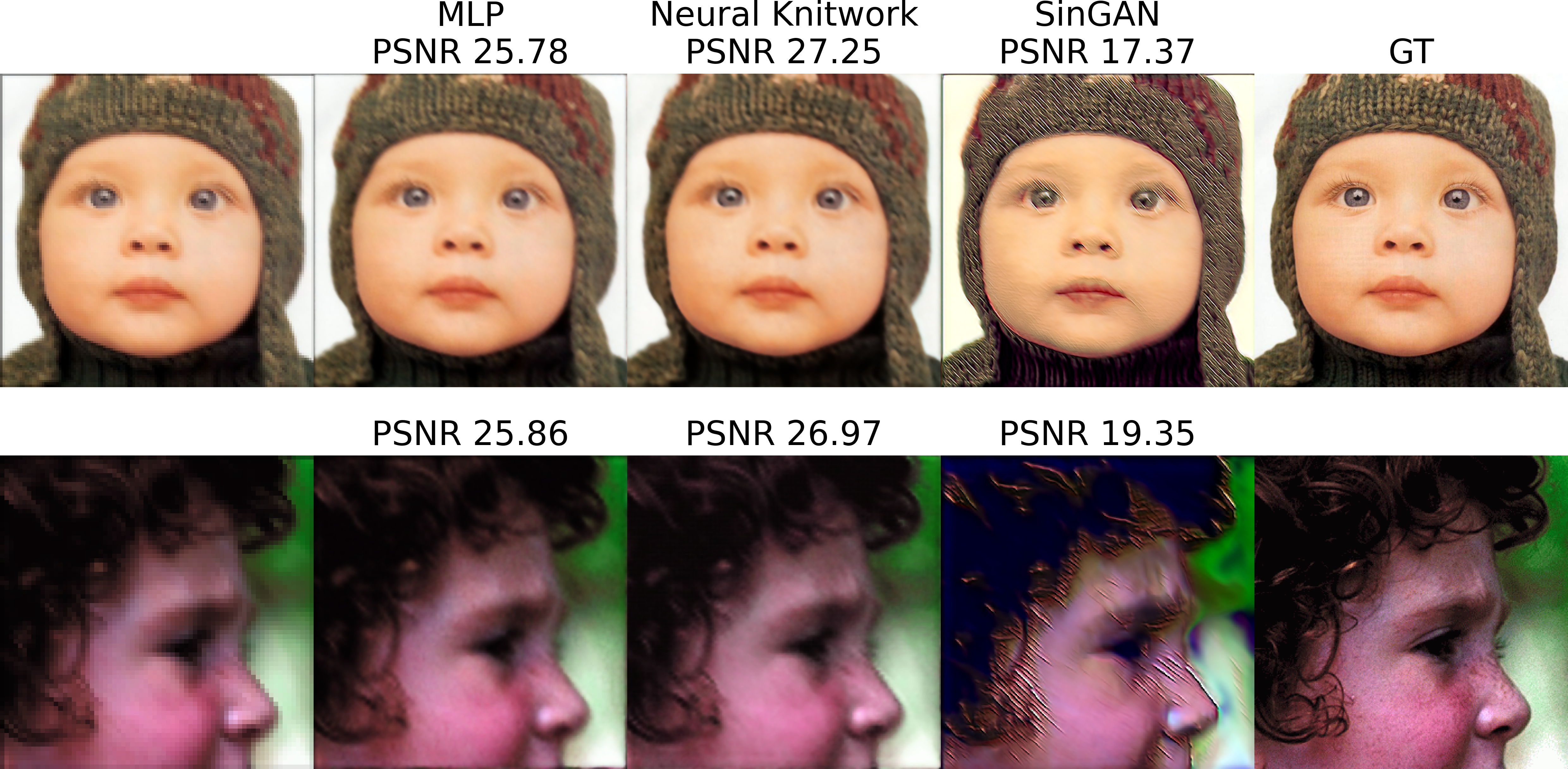}
        \caption{Comparison of blind image super-resolution for a diagonal Gaussian kernel and upscaling factor of 4x. Neural Knitwork can outperform conventional coordinate \gls{mlp} network and achieve higher PSNR. SinGAN, while generating a considerable amount of high frequency details, results in significant artifacts.}
        \label{fig:blind_sr_images}
    \end{figure*}

    Figure~\ref{fig:blind_sr_images} contains results for a diagonal kernel and upscaling factor of 4, for the proposed Neural Knitwork, the conventional \gls{mlp} and SinGAN, another image super-resolution method based on internal learning. The results show that SinGAN has the lowest performance in terms of PSNR but it also creates distinguishable artifacts. Table~\ref{tab:sr_comparison} shows how Neural Knitwork compares to counterparts along with the model sizes. Interpolation with conventional \gls{mlp} directly implies delta kernel and hence, they perform best in this instance. For other kernels, a Neural Knitwork can boost the performance in some instances by adjusting to the kernel.
    
    \begin{table}[]
        \centering
        \caption{We compare the blind super-resolution performance achieved by a conventional coordinate \gls{mlp}, a \gls{cnn}-based internal learning framework of SinGAN and our method. We compute PSNR ($\uparrow$) and SSIM ($\uparrow$) for a number of upscaling factors and downsampling kernels.}
        \begin{tabular}{cclccc}
        \hline
            Dataset & Factor & Kernel & MLP & SinGAN~\cite{Shaham2019} & Neural Knitwork (ours) \\
            \hline\hline
             & & Delta & \textbf{31.58 / 0.95} & 19.22 / 0.65 & 27.39/0.88 \\
            Set5 & 2$\times$ & Diagonal Gaussian  & 23.78 / 0.83 & 19.95 / 0.72 & \textbf{24.62 / 0.82} \\
             & & Round Gaussian & 24.95 / 0.86 & 21.59 / 0.75 & \textbf{25.48 / 0.84} \\
            \hline
             & & Delta & \textbf{25.38 / 0.85} & 17.16 / 0.53 & 23.81 / 0.81 \\
            Set5 & 4$\times$ & Diagonal Gaussian & 23.47 / 0.81 & 19.15 / 0.66 & \textbf{24.22 / 0.82} \\
             & & Round Gaussian & 24.61 / 0.84 & 20.75 / 0.72 & \textbf{25.36/0.84} \\\hline
             & & Delta & \textbf{27.22 / 0.89} & 14.21 / 0.41 & 24.31/0.82 \\
            Set14 & 2$\times$ & Diagonal Gaussian & \textbf{22.09 / 0.75} & 16.96 / 0.56 & 22.08 / 0.74 \\
             & & Round Gaussian & \textbf{22.96 / 0.78} & 17.21 / 0.57 & 22.35/0.75 \\
            \hline
             && Delta  & \textbf{22.45 / 0.76} & 14.32 / 0.33 & 21.72/0.75 \\
            Set14 & 4$\times$ & Diagonal Gaussian  & 21.90 / 0.73 & 17.75 / 0.56 & \textbf{22.05/0.73} \\
             & & Round Gaussian& \textbf{22.52 / 0.76} & 18.65 / 0.62 & 21.56/0.71 \\
            \hline
            \rule{0pt}{2.5ex} Parameters & & & 263K & 2,381K & 608K \\
        \end{tabular}
        \label{tab:sr_comparison}
    \end{table}

    \subsection{Denoising}
    
\begin{figure*}[b]
        \centering
        \includegraphics[width=\textwidth]{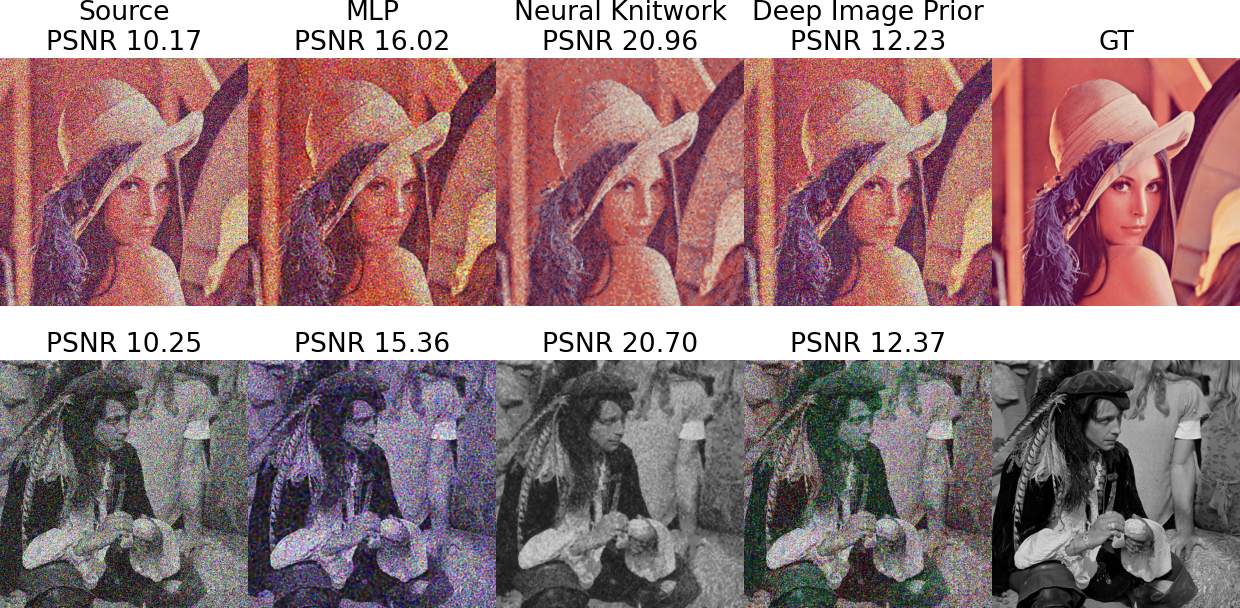}
        \caption{Neural Knitwork demonstrates superior performance for severe levels of noise, in this case $\sigma = 40$.}
        \label{fig:denoising_images}
    \end{figure*}

    As we demonstrate in Figure~\ref{fig:denoising_images}, a standard \gls{mlp} network has limited denoising capability because it attempts to fit all pixel colors with no additional constraints. In contrast, a Neural Knitwork ensures that both patches and pixel colors are reliably reconstructed while imposing additional consistency constraint on the derived solution. In the illustrated result with severe noise levels of $\sigma$ = 40, we achieve PSNR approximately 4 dB higher than in the case of a conventional coordinate \gls{mlp}. Further, Table~\ref{tab:denoising_comparison} confirms that the Neural Knitwork model outperforms both other methods for high noise levels.
    
    \begin{table}[]
        \centering
        \caption{Comparison of achieved denoising performance. For higher power levels, the Neural Knitwork achieves higher PSNR ($\uparrow$) and SSIM ($\uparrow$) than a conventional \gls{mlp} and \gls{dip}.}
        \begin{tabular}{ccccc}
        \hline
            Dataset & $\sigma$ & MLP & \gls{dip}~\cite{Ulyanov2020} & Neural Knitwork (ours) \\
            \hline
            \hline
             & 10 & 23.58/0.70 & \textbf{27.56/0.85} & 26.69/0.83 \\
            Set5 & 20& 17.76/0.42 & 19.15/0.49 & \textbf{21.69/0.63} \\
             & 40 & 12.43/0.19 & 11.6/0.15 & \textbf{15.91/0.37} \\
            \hline
             & 10 & 24.94/0.77 & \textbf{26.95/0.84} & 26.15/0.85 \\
            Set14 & 20 & 19.56/0.55 & 19.70/0.53 & \textbf{23.08/0.73} \\
             & 40 & 14.55/0.30 & 12.62/0.19 & \textbf{18.75/0.56} \\
            \hline
            \rule{0pt}{2.5ex} Parameters & & 263K & 2,400K & 512K \\
        \end{tabular}
        \label{tab:denoising_comparison}
    \end{table}

\section{Conclusion}

    Neural Knitworks constitute a hybrid architectural approach for internal learning applications, based on three shallow \gls{mlp} networks. It enhances conventional coordinate-based \gls{mlp} networks by adding synthetic capabilities for tasks such as inpainting, super-resolution, and denoising, at levels comparable or better than the considered alternatives. Furthermore, the Neural Knitwork used in our experiments is 5x~smaller than \gls{cnn} internal learning counterparts with an additional benefit of being fully parallelizable; that is, all coordinate outputs could be computed independently. Apart from the significant potential for speed up, Neural Knitworks have the advantage of precise control over the output image size by adjusting the set of input coordinates. Our experimentation shows that Neural Knitworks can be sensitive to hyperparameters such as individual loss weights, patch sizes, and learning rates, however, the configuration used in our experiments has shown to offer stable performance.

{\small
\bibliographystyle{ieeetr}
\bibliography{egbib}
}

\end{document}